\documentclass[journal]{IEEEtran}

\usepackage{enumerate}
\usepackage{graphicx}
\usepackage{subfig}
\usepackage[ruled,vlined]{algorithm2e}
\usepackage{algorithmic}
\usepackage{amsfonts,amssymb,amsmath}
\usepackage{dsfont}
\usepackage{multirow}
\usepackage[table]{xcolor}
\usepackage{colortbl}
\newcommand{\tabincell}[2]{\begin{tabular}{@{}#1@{}}#2\end{tabular}}
\DeclareMathOperator*{\argmax}{arg\,max}
\DeclareMathOperator*{\argmin}{arg\,min}

\begin{document}
\title{Introducing Memory and Association Mechanism Into a Biologically Inspired Visual Model}
\author{Hong~Qiao,~\IEEEmembership{Senior~Member,~IEEE,}
        Yinlin~Li,~\IEEEmembership{Student~Member,~IEEE,}
        ~Tang~Tang,~\IEEEmembership{Student~Member,~IEEE}
        and~Peng~Wang,

\thanks{Manuscript received July 3, 2013; revised.}
\thanks{H. Qiao is with the Institute of Automation, Chinese Academy of Sciences,
Beijing 100190, China (e-mail: hong.qiao@ia.ac.cn).}
\thanks{Y.L. Li and T. Tang are with the Graduate School and the State Key Laboratory of Management
and Control for Complex Systems, Institute of Automation, Chinese Academy of Sciences.}}


\maketitle

\begin{abstract}
A famous biologically inspired hierarchical model firstly proposed by Riesenhuber and Poggio has been successfully applied to multiple visual recognition tasks.  The model is able to achieve a set of position- and scale-tolerant recognition, which is a central problem in pattern recognition.

In this paper, based on some other biological experimental results, we introduce the Memory and Association Mechanisms into the above biologically inspired model.  The main motivations of the work are (a) to mimic the active memory and association mechanism and add the 'top down' adjustment to the above biologically inspired hierarchical model and (b) to build up an algorithm which can save the space and keep a good recognition performance.

More details of the work could be explained as follows:

\begin{enumerate}[(1)]
\item
In objects memorizing process: Our proposed model mimics some characteristics of human's memory mechanism as follows:

    (a)\hspace{10pt} In our model, one object is memorized by semantic attributes and special image patches (corresponding to episodic memory).   The semantic attributes describe each part of the object with clear physical meaning, for example, if eyes and mouths of faces are 'big' or 'small' and so on.  One special patch is selected if the value of the corresponding semantic feature is far from average one.  The patch should be the most prominent part of the object.

    (b)\hspace{10pt} In our model, different features (semantic attributes and special patches) of one object are stored in distributed places and the common feature of different objects is saved aggregately, which can learn to classify the difference of similar features of different objects.  The similarity thresholds to each object can be learnt when new objects are learnt.
\item
In object recognition process: In biological process, the associated recognition including familiarity discrimination and recollective matching.   In our proposed model, firstly mimicking familiarity discrimination ('knowing' though the episode), we compare the special patches of candidates with that of saved objects using above mentioned biologically inspired hierarchical model, where the candidates and saved objects have the same prominent semantic features.  Then mimicking recollective matching, the comparison results of special patches are combined with semantic feature comparison.
\end{enumerate}

The new model is also applied to object recognition processes.  The primary experimental results show that our method is efficient with much less memory requirement.
\end{abstract}


\IEEEpeerreviewmaketitle

\section{Introduction}\label{Section1}
\IEEEPARstart{M}{any} researchers have established a series of neural computational models for vision processes based on the biological mechanism and showed that the models can be well applied to pattern recognition
\cite{HubelReceptive,RiesenhuberHierarchical,IttiAModel,RustHowMT}.  More importantly, recently, biological research with information technology integration is an important trend of research in all over the world.

In particular, in 1999, Riesenhuber and Poggio proposed a famous neural computational model for vision process \cite{RiesenhuberHierarchical}. This model is a hierarchical feed-forward model of the ventral stream of primate visual cortex, which is briefly called as HMAX. The model is closely related to biological results.   Each level in the model was designed according to data of anatomy and physiology experiments, which mimicked the architecture and response characteristics of the visual cortex.

Furthermore, Giese and Poggio have extended the above model to biological motion recognition \cite{GieseNeural}.  They established a hierarchical neural model with two parallel process streams, i.e. form pathway and motion pathway. A series of extended models have given very good performance in biological motion recognition in cluster \cite{sigala2005learning} and are compared with psychophysics results \cite{CasileCritical}.  The model was also extended to a computational model for general object recognition tasks \cite{SerreRobust}, which can output a set of position and scale invariance features by alternating between a template matching and a maximum pooling operation corresponding to S1-C2 layers.  Some other researchers introduced sparsification, feedback and lateral inhibition \cite{HuangEnhanced,mutch2006multiclass}into HMAX.  The series works have demonstrated very good performance in a range of recognition tasks which is competitive with the state of art approaches, such as face recognition \cite{MeyersUsing}, scene classification \cite{HuangBiologically} and handwritten digit recognition \cite{HamidiInvariance}.

From another point of view, Itti established a visual attention model \cite{IttiAModel,IttiComputational}, which is inspired by the behavioral and neural architecture of the early primate visual system.  Itti and Poggio et al \cite{walther2002attentional} merged the saliency-based attention model proposed in the previous work \cite{IttiAModel} to HMAX in order to modify the activity of the S2 layer.  Here the S2 layer mimics the response characters of V4 in the primate visual cortex, which shows an attention modulation in electrophysiology and psychophysics experiments.

Saliency based visual attention models are further compared with behavioral experiments \cite{PetersApplying} and showed that the models could account for a significant portion of human gaze behavior in a naturalistic, interactive setting. A series of visual attention models \cite{BorjiState} have demonstrated successful applications in computer vision \cite{MaAGeneric}, mobile robotics \cite{lang2003providing}, and cognitive systems \cite{FrintropComputational}.

Some other work has tried to combine HMAX and Deep Belief Networks.  The HMAX model could extract position and scale invariance features with a feed-forward hierarchical architecture, but the lack of feedback limits its performance in pure classification tasks.  Deep Belief Networks (DBN) \cite{HintonAfast} has shown the state of the art performance in a range of recognition tasks \cite{lee2009convolutional,MohamedAcoustic}.  DBN uses generative learning algorithms, utilizes feedback at all levels and provides ability to reconstruct complete or sample sensory data, but DBN lacks the position and scale invariance in the HMAX. Therefore, combing DBN with the HMAX may be meaningful to extract more accurate features for pattern recognition.

The above models are all biologically inspired models and have been successfully applied in practical work.  In this paper, based on other biological results, we try to introduce the Memory and Association into the HMAX.  Our method can reduce memory requirement and achieve comparable accuracy to the state of art approaches.



\section{Related Works}\label{Section2}
In this paper, we propose a model based on HMAX and some basic facts about the memory and association mechanism established over the last decade by several physiological studies of cortex \cite{TulvingEpisodic,MartinSemantic,SquireRecognition,McElreeIsolating}. The accumulated evidences points are summarized as follows.
\\
\begin{enumerate}[(1).]
\item
\textit{The Memory About an Object Includes the Episodic and Semantic Memory}

Tulving et al. proposed that episodic memory and semantic memory are two subsystems of declarative memory \cite{McElreeIsolating}.  Though the two systems share many features, semantic memory is considered as the basis of episodic memory which has additional capabilities that semantic memory does not have. In the memory encoding process, information is encoded into semantic memory first and can then be encoded into episodic memory through semantic memory, as shown in Fig. \ref{TwoMemories}.
\\

\begin{figure}[h]
\begin{center}
\includegraphics{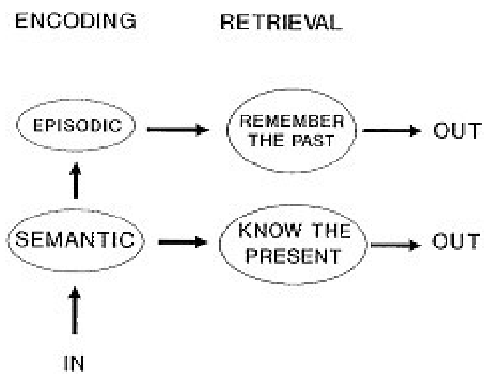}
\end{center}
\caption{Sketch of the relationship between semantic and episodic memory \cite{TulvingEpisodic}}
\label{TwoMemories}
\end{figure}

\item
\textit{An Object Concept May be Represented by Discrete Cortical Regions}

Brain imaging studies show that object representation is related to different cortical regions, forming a distributed network that is parallel to the organization of sensory and motor systems \cite{MartinSemantic}. For example, in the word generation experiments, ventral and lateral regions of the posterior temporal cortex have been found eliciting differentially with different types of information retrieved \cite{MartinDiscrete}.  In other related behavioral experiments in which the subjects are required to name an action or a color, the specialization of different cortical regions has also been observed \cite{MartinDiscrete}.
\\

\item
\textit{Neurons Responding to the Common Feature of Different Objects Aggregate Together}

Researchers believe that the brain tends to process common features from different objects in the same areas \cite{MartinSemantic}. One supporting fact for the assertion is the existence of the cortical region responding to the shape attributes of different visual categories.  In the related behavioral experiments, when the subject was required to perform same task with different object categories as stimuli, the ventral occipitotemporal cortex was consistently activated and encoded the object forms with distinct neural response patterns \cite{MartinSemantic}.

Another body of evidence comes from studies of cortical lesions \cite{HartDelination,HodgesSemantic}. The damage to temporal lobes is found to be strongly associated with the impairment of abilities to recognize objects and retrieve information about object-specific characteristics \cite{HartDelination,HodgesSemantic}. This implies that the temporal lobe is the region in which object-specific information from all categories is commonly stored.

Functional column may provide the anatomical basis accounting for the above phenomena, and recent studies suggest that the functional column is a basic processing architecture spreading over visual neural systems \cite{WangOptical,FujitaColumns,StrykerNeurobiology}. On the surface of infratemporal cortex that activated by faces, neurons with common preference were found aggregating in patches specifying different stimuli \cite{WangOptical}.

Similar organization was also seen in the area of posterior TE that is responsive to simple two-dimensional shapes \cite{FujitaColumns}. It is inferred that such a columnar organization would produce stimulus invariance \cite{ToveeIsFace} and also provides a visual alphabet from which numerous complex objects can be efficiently represented as a distributed code \cite{StrykerNeurobiology}.
\\

\item
\textit{Recognition Memory Includes Familiarity and Recollection Components}

As a basic functional part of memory, recognition memory has its use in identifying and judging whether the presented object has ever been captured consciously \cite{BrownRecognition,SquireRecognition}. It is widely accepted that two components make up the recognition memory:  One is the familiarity discrimination that determines 'knowing' though the episode in which the object has appeared may not be recalled, and the other is the recollective matching that means remembering both the presence of the object and the related episode.

Different mechanisms have been proposed to explain the two recognition experiences. Some researchers argue that the only difference between familiarity discrimination and recollection matching is the trace strength, in term of which the former is weaker than the latter \cite{HaistOn,HirshmanModeling,DonaldsonThe}, while others regard the two processes as qualitatively different modes of recognition memory, which may be executed by hippocampus and perirhinal cortex, respectively \cite{BrownRecognition,SquireRecognition}.
\\

\item
\textit{Familiarity Recognition is Rapid and Accurate and Only Needs a Small Number of Neurons}

From the evolutionary perspective, owning a simplified recognition system focusing on familiarity is likely to have advantage in speed of response to novelty by saving time of deep recalling. This hypothesis has been confirmed in some related experiments, in which subjects made familiarity decisions faster than recollect decisions \cite{SeeckEvidence, HintzmanRetrieval,McElreeIsolating}.

The computational efficiency of a recognition system dedicated for familiarity discrimination has also been demonstrated by the simulated neural networks \cite{bogaczhigh}. Compared with systems relying on associative learning, familiarity decision can be made accurate and faster by a specially designed network with smaller size of neuron population.
\\
\end{enumerate}

All of above works form the basis of our model which would be explained in Section III.

\section{The New Model-Introduction Memory and Association into Biologically Inspired Model}\label{Section3}
\begin{figure*}[!t]
\begin{center}
\includegraphics[width=0.6\textwidth]{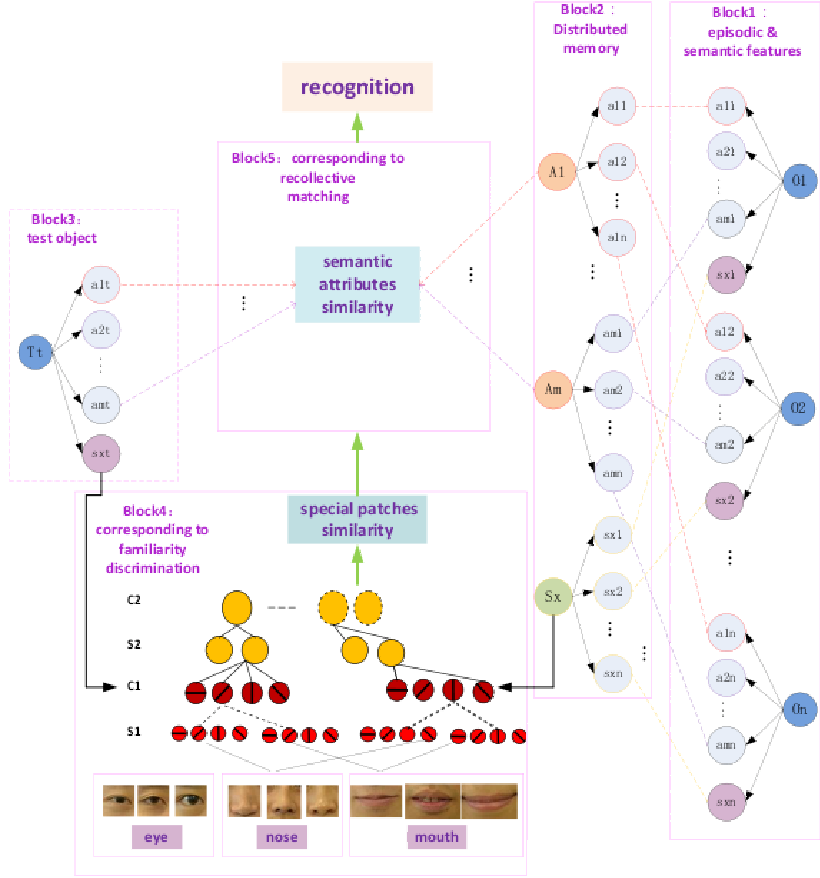}
\end{center}
\caption{The model proposed - Introducing biologically inspired memory and association into HMAX model.}
\label{OurModel}
\end{figure*}

The new model is presented in Fig. \ref{OurModel}.  This section includes 5 parts, which introduce the framework of our model and algorithms related to sub processes.

\subsection{The Framework of Our Model}
Figure III1 presents our model which introduces biologically inspired memory and association model into HMAX.  Our framework consists of 5parts, i.e., block 1 to block 5.
\begin{enumerate}
\item\textbf{Objects are Memorized Through Episodic and Semantic Features (Block1)}

  $O_i$ represents the $i$th object $(i=1,\ldots n)$, $a_ji$ represents the $j$th semantic attribute $(j=1,\ldots m)$ of the $i$th object, and $s_{xi}$ represents the $x$th special episodic patch of the $i$th object.  In face recognition process, semantic description can be 'the eyes are large', 'the month is small' and so on.  The special episodic patch can be the image of an 'eye' if the eyes are prominent on the face.  These two kinds of features are just corresponding to the semantic and episodic memory of declarative memory in cognitive science \cite{TulvingEpisodic}

\item\textbf{Features of One Object are Saved in Distributed Memory and of and Common Features of Various Objects Are Stored Aggregately (Block2)}

  $A_1$ to $A_m$ represent various common semantic attributes of different objects. $S_x$ is a library for special episodic patches of different objects.  A common attribute of different objects is stored in the same regions, for example, $a_{11}-a_{1n}$ is the first feature of different objects $(i=1,\ldots n)$ which are stored together, called as $A_1$. $A_1$ not only has clear biological area but also has learning ability. For example, the similarity sensitivity would be learnt when new objects are saved. This memory mechanism has clear biological evidences \cite{ToveeIsFace, MartinSemantic}

\item\textbf{Recognition of One Candidate (Block3)}

  $T_t$ represents a candidate. The semantic features $a_{1t}$ to $a_{mt}$ and episodic patch feature $s_{xt}$ are extracted before recognition.

\item\textbf{Familiarity Discrimination (Block4)}

Familiarity discrimination is achieved through the HMAX model.  Both $S_x$ and $s_{xt}$ are extracted in the C1 layer of HMAX model.  The saved object can be ordered by comparing similarity between the candidate and the saved objects in the C2 layer of HMAX model. This process corresponds to the familiarity discrimination ('knowing' through the episode) in biological recognition memory \cite{BrownRecognition}.

\item\textbf{Recollective matching (Block5)}

Recollective matching is achieved through integration of semantic feature and episode feature similarity analysis.  We compare the semantic features of the candidate with those of the top saved objects according to familiarity discrimination. If the difference between the candidate and the closest object does not exceed the threshold which we have learnt during the memory process, we consider that the candidate 'is' the object. This procedure is illuminated by recollective matching in recognition memory \cite{BrownRecognition}.

\end{enumerate}

\subsection{Encoding Episodic and Semantic Features (Block 1)}
As we know, memory is the process by which information is encoded, stored, and retrieved, and the memory to an object with conscious process is called declarative memory, which can be divided into semantic memory and episodic memory \cite{TulvingEpisodic, SquireEpisodic}.  Semantic memory refers to the memory which is learned by people, such as meanings, understandings, concept-based knowledge and work-related skills and facts \cite{SaumierSemantic}.   Episodic memory refers to the memory which is explicitly located in the past involving the specific events and situations, such as times, places, associated emotions and other contextual knowledge \cite{EichenbaumACortival, ConwayEpisodic}.

\begin{figure}[h]
\begin{center}
\includegraphics[width=\columnwidth]{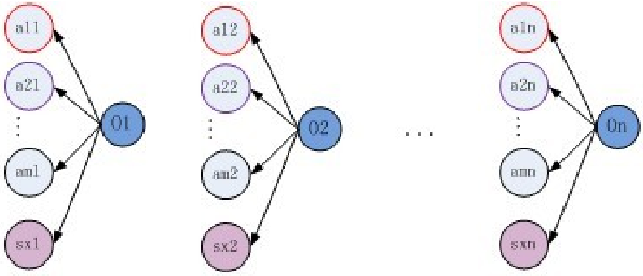}
\end{center}
\caption{Semantic attributes and special episodic patches in memory.}
\label{Semantic}
\end{figure}

In this paper, the features of a saved object $i$ include descriptive attributes $a_{1i}$ to $a_{mi}$ and special patch $s_{xi}$ (Fig. \ref{Semantic}), where descriptive attributes correspond to semantic features, and special patches correspond to episodic features.

\subsubsection{Semantic Features}\label{SectionSemantic}
In biological process, human would memorize an object using semantic features. In computational process, semantic features can reduce the memory size requirement. However, in order to get semantic features, a big dataset and learning ability are needed, which correspond to the prior knowledge.

A simplified algorithm is presented as follows.  First, we need to extract patches $g_{ji}$ with clear physical meaning (such as the images of eyes, months of one face) for each object $i$.  Then, we can get an average view $\bar{g}_j$ of each feature $g_{ji}$.  At last, we compare $g_{ji}$ and  $\bar{g}_j(j=1,\ldots m)$ to obtain the semantic attribute $a_{ji}$ of a particular object. The detailed algorithm is given in Algorithm \ref{Alg1}.
\begin{algorithm}[t]
    \caption{Extracting semantic attributes.}
    \label{Alg1}
    \SetKwInOut{Input}{Input}
    \Input{objects $o_i, i=1\ldots n$\\
    features $g_{ji}, j=1\ldots m$}

    \begin{algorithmic}[1]
    \STATE Extract geometric features $(j=1,\ldots m)$ from object $o_i$
    \STATE Compute a common view for each feature ${\bar{g}_j=f(g_{j1},g_{j2},\ldots g_{jn})}$
    \STATE Compute semantic attributes $a_{ji}=f(g_{ji},\bar{g}_j)$
    \end{algorithmic}
    \SetKwInOut{Output}{Output}
    \Output{Semantic attributes $a_{ji}$.}
\end{algorithm}

Take face recognition as an example. In this paper, the active shape model (ASM) is used to extract typical points from faces. Avoiding to describe one part 'big' or 'small', which can be vague, in this paper, we compute effective geometric features of each part related declarative features of eyes, nose and mouth (Table \ref{GeoFeature}).  We could establish an average value for each geometric feature.  By comparing each individual with the average face, we can get the computational semantic attributes for each part.

\begin{table}[h]
\centering
\begin{tabular}{|c|c|c|c|c|c|}
\hline
eyes&    \tabincell{c}{eye\_area\\eye\_length\\eye\_height\\eye\_ratio(l/h)\\brow\_eye\_distances\\eye\_cheek\_ratio}&
nose&
\tabincell{c}{nose\_width\\nose\_ratio\\court\_ratio}&
mouth&
\tabincell{c}{mouth\_length\\mouth\_height\\mouth\_ratio\\Philtrum\_length}\\
\hline
\end{tabular}
\caption{15 geometric features of a face.}
\label{GeoFeature}
\end{table}

In experiments, we would show that these geometric features have coincidence with general semantic meaning, such as 'the eyes are large' and so on.

It should be noted that representing the geometric features of a face component should consider the influences of expression, view direction, light and even subjective feeling of an observer.  In order to get fair results, our current dataset is collected with frontal faces with less expression, and the geometric attributes are normalized as $g_{ji}=\frac{g_{ji}-u_j}{\sigma_j}$, where $u_j$ is the average of the $j$th geometric feature and $\sigma_j$ is the standard deviation of the $j$th  geometric feature.

For one person, many other attributes, such as gender, race and age can be included in further research. Some events and their relationships can enrich the memory processes.

\subsubsection{Episodic Patches}
In order to mimic episodic memory, we need to find out the dominant patches of an object through finding out the most prominent semantic features  of the object I (Equation 3-2).

Take face recognition as an example again.  First, we need to find out the most prominent semantic feature for physical parts, for example, eyes, nose and mouth:

\begin{equation}
J_i=\argmax\limits_{j=1,2,\ldots m}(\|a_{1i}-\bar{a}_{2}\|_2,\|a_{2i}-\bar{a}_{1}\|_2,\dots \|a_{mi}-\bar{a}_{m}\|_2)
\end{equation}

where $J_i$ is the dominant semantic feature of the $i$th object, $a_ji$ is the $j$th semantic feature of the $i$th object, and $\bar{a}_{ji}$ is the $j$th average semantic feature.
Special episodic patches could be extracted, which corresponds to $J_i$  of the object $O_i$ and are put into HMAX model proposed by \cite{RiesenhuberHierarchical}, which can extract a set of position and scale tolerant features.

Different from the original HMAX model, only episodic patches corresponding to the prominent part of a candidate are put into the model, and only those of the 'known' objects are stored in a library.

\subsection{Distributed Memory Structures (Block 2)}
How to remember and organize different features is a key problem in the memory process. It would also influence the retrieval and association processes. This is also an important part of our framework.

The studies of cognitive sciences have indicated that the extraction, storage and retrieval of different features are realized by the distributed cortical areas. As we have listed in section \ref{Section2}, many data and evidence proved that an object concept may be represented by the discrete cortical regions \cite{MartinDiscrete} and neurons responding to the common feature of different objects aggregate together \cite{ToveeIsFace}.

\begin{figure}[h]
\begin{center}
\includegraphics[width=\columnwidth]{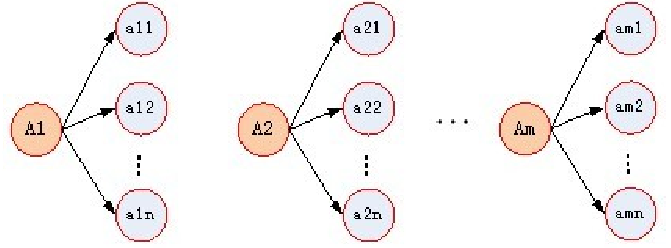}
\end{center}
\caption{Distributed memory structure.}
\label{DisMemo}
\end{figure}

In this paper, we mimic this distributed structure to memorize and retrieve features. Suppose there are $m$ visual cortical regions $A_1$ to $A_m$(Fig. \ref{DisMemo}), which are sensitive to different kinds of features $a_{1,t}$ to $a_{m,t}$. Thus candidate which have feature $a_{i,t}$ will active a distinctive response in region $A_i$.

This kind of distributed structures has great advantages.  The visual cortical region $A_i(i=1, \ldots, m)$ can be more and more sensitive and effective through comparing different $a_{i,t}$.  Two similarity thresholds for semantic attributes and episodic patches which would decide if a candidate is a 'known' object are learned when a new object is memorized.  In the further work, the attributes of one object would also be connected and influence each other.

\subsection{Recognition based on Familiarity Discrimination and Recollective Matching (Block 3)}
As shown in Section II, there are two processes in the recognition memory: familiarity discrimination and recollective matching \cite{BrownRecognition}, and their combination is useful for a fast and accurate human recognition task.  In the proposed framework, we also have two processes, corresponding to familiarity discrimination and recollective matching.
\subsubsection{Familiarity Discrimination}
In block 1, we established a library of episodic patches $\{S_x\}$ of different 'known' objects in the memory process. In the familiarity discrimination of the association process, a special episodic patch $s_{xt}$ of the candidate $T_t$ is extracted and put into the C1 layer of the HMAX model.  Then we compute the S2 and C2 level features of the candidate and the 'known' objects which have the same prominent semantic attribute.  Finally, we sort the 'known' objects through their C2 feature similarity compared with the candidate.

This process corresponds to the familiarity discrimination in recognition memory \cite{ BrownRecognition}.

\subsubsection{Recollective Matching}
We compare the semantic features $a_{i,t}(i=1,\ldots,m)$  of the candidate with those of the top 'known' objects obtained from familiarity discrimination.  If the smallest dissimilarity does not exceed the thresholds which we have learnt during memory process, the candidate is 'recognized' as the closest 'known' object.  Otherwise, we regard the candidate as a new one. This procedure is illuminated by re-collective matching in recognition memory \cite{ BrownRecognition}.

The whole algorithm is given as algorithm \ref{Alg2}:

\begin{algorithm}[h]
    \SetKwInOut{Memo}{memory process}
    \SetKwInOut{Rtrv}{Retrieval Process}
    \Memo{semantic attributes, episodic patch features, dominant attributes, and similarity thresholds of the training objects}
    \Rtrv{A candidate}
    \caption{Memory and retrieval process}

    \begin{algorithmic}[1]
    \STATE Extracting semantic attributes, dominant attributes and the episodic patch
    \STATE \textbf{Familiarity Discrimination:} finding the training objects which have the same prominent attribute $j$, and using their stored patches library $S_j$ to get the C2 layer features for both $f^{C2}_{test}$ and $f^{C2}_i$, and sort the candidates by $\|f^{C2}_{test}-f^{C2}_i\|_2$
    \STATE \textbf{Recollective Matching:} comparing the detailed semantic features of the test object with those of the top candidate training objects
    \[
    similarity_i=\argmin_{i\in top3}\sum^m_{j=1}(\|a_{jtest}-a_{ji}\|_2)
    \]
    \STATE \textbf{Zero-shooting object memorizing:} Store feature set of the test object
    \end{algorithmic}
    \label{Alg2}
\end{algorithm}

\section{Experiment}
\begin{figure}[h]
\begin{center}
\includegraphics[width=\columnwidth]{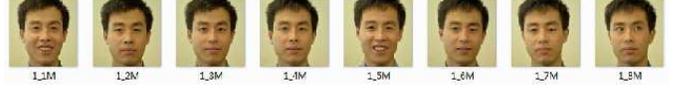}
\end{center}
\caption{The first person's 8 pictures used in the memory process.}
\label{SAMP1}
\end{figure}

In this section, we conduct experiments to demonstrate the effectiveness of our model through introducing biologically inspired memory and association mechanism into HMAX.
We first establish a dataset which consists of front faces of 7 persons (Fig. \ref{IDENT}).  The first 5 persons' face images are used in both the memory and association phases and other 2 persons' face images are only used in the recognition phase.  Some of the samples in the dataset are shown in Fig. \ref{SAMP1}:
\begin{figure}[t]
\begin{center}
\includegraphics[width=\columnwidth]{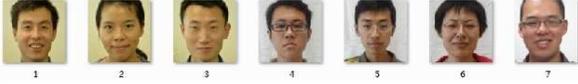}
\end{center}
\caption{7 persons whose pictures are used in experiments.}
\label{IDENT}
\end{figure}

The experimental details are given as follows.

\subsection{Special Patches Evaluation}
\begin{figure}[h]
\begin{center}
\includegraphics[width=\columnwidth]{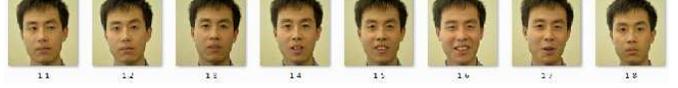}
\end{center}
\caption{The first person's 8 candidate images.}
\label{SAMP2}
\end{figure}

The first person's 8 candidate images are presented in Fig. \ref{SAMP2}.
Responses of the C2 units in HMAX model are used for the dissimilarity analysis of special patches.

Although light, expression, scales and other factors would influence the similarity analysis, the experimental results show that the similarity of the special patches through HAMX primarily matches the semantic description.  For example, one candidate's special patch is more similar to that with the same semantic description, that is, the big eyes have high similarity with big eyes and low similarity with small eyes.

\begin{figure}[h]
\begin{center}
\includegraphics[width=\columnwidth]{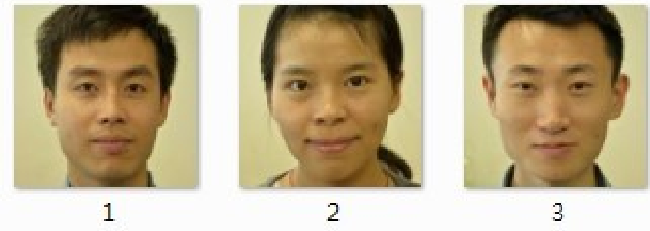}
\end{center}
\caption{The first and second persons have with 'big' eyes and the third one has with small eyes.}
\label{Eyes}
\end{figure}

In the memory process, the special patches of the first 3 persons' images are all eyes.  The eyes images are scaled with 4 and 6, and the first 2 with 'big' eyes and the third one with 'small' eyes (Fig. \ref{Eyes}).

\begin{figure}[h]
\begin{center}
\includegraphics[width=\columnwidth]{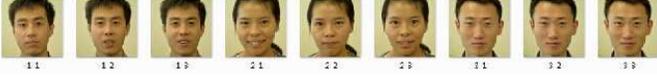}
\end{center}
\caption{Candidates in Special Patch Evaluation.}
\label{Cand}
\end{figure}

In recognition process, three other pictures of the first 3 persons are used(see Fig. \ref{Cand}).

The similarity analysis and rank are presented in Table \ref{SIMI}.

\begin{table}[h]
\begin{center}
\begin{tabular}{|c|c|c|c|c|}
\hline
\multirow{2}*{patch 1.4}    &C2 value   &0.0088 &0.0096 &0.0136\\
\cline{2-5}
                            &Ranking	&2     &1      &3\\
\hline
\multirow{2}*{patch 2.4}    &C2 value   &0.0032 &0.0113 &0.0165\\
\cline{2-5}
                            &Ranking	&2     &1      &3\\
\hline
\multirow{2}*{patch 3.4}    &C2 value   &0.0028 &0.0088 &0.0117\\
\cline{2-5}
                            &Ranking	&3     &2      &1\\
\hline
\multirow{2}*{patch 1.6}    &C2 value   &0.0205 &0.0300 &0.0359\\
\cline{2-5}
                            &Ranking	&1     &2      &3\\
\hline
\multirow{2}*{patch 2.6}    &C2 value   &0.0080 &0.0380 &0.0472\\
\cline{2-5}
                            &Ranking	&2     &1      &3\\
\hline
\multirow{2}*{patch 3.6}    &C2 value   &0.0055 &0.0285 &0.0438\\
\cline{2-5}
                            &Ranking	&3     &2      &1\\
\hline
\end{tabular}
\end{center}
\caption{similarity analysis and rank.}
\label{SIMI}
\end{table}

Where, $Patch_{ij}$ represents an eye patch with scale $j\times j$  of the $i$th person. The patches are ranked according to their average values from the C2 layer.

It is not surprising that the eye patches of the first two persons generate higher values with portraits having bigger eyes, while the patches of the third person prefer the smaller ones. This rule always holds even in some cases that the eyes of the first two identities are confused.

The results above demonstrate that the low-level features used in the model are reliable indicators of visual properties. By combining a sufficient number of such features, visual objects can be efficiently represented by the response patterns, just like the neural activities in certain cortical regions by which they are encoded.

\subsection{Familiarity Discrimination}
The purpose of familiarity discrimination is to 'feel' if 'knowing' the subject.  After a special patch similarity analysis, it is necessary to decide if the subject is 'known' rapidly.  Then the problem is how to design the threshold of each saved subject which can be used to decide if the candidate is the saved subject.

In this paper, the threshold for a 'known' face image i is given as follows,

\begin{eqnarray}
\begin{split}
&&thres_{1i}&=\argmax_i(\|f^q_i-\bar{f}_i\|_2)\\
&&thres_{2i}&=\argmax_{j,j\neq i}(\|f^q_i-\bar{f}_j\|_2)\\
&&thres\ \ &=thres_{1i}+\|thres_{2i}-thres_{1i}\|/\lambda\\
\end{split}
\end{eqnarray}

where $f$ represents the semantic features or C2 features of special patches, $f_i^q$ is the $q$th image feature of the $i$th object ($i$th object has $q$ images), $\bar{f}_i$ is the average features of $q$ images belonging to $i$th object, $thres_{1i}$ is the intra-class maximum difference, $thres_{2i}$ is the inter-class minimum difference, $\lambda$ is the ratio of $thres_{1i}$ and $thres_{2i}$ and $thres$ is the final similarity threshold.

Clearly, the larger the threshold is, the more insensitive the image is. $\|thres_{2i}-thres_{1i}\|/\lambda$ is used to adjust the $thres_{1i}$ to make it more flexible.

The thresholds for 5 persons' images are shown in following Table \ref{THRES}:

\begin{table}[h]
\begin{center}
\begin{tabular}{|c|c|c|c|c|c|}
\hline
        &\multicolumn{3}{c|}{mouth} &\multicolumn{2}{c|}{Eyes}\\
        \hline
identity    &1 &2  &4  &3  &5\\
\hline
patch\_thres &0.5406 &0.5203 &1.5492 &0.4174 &0.4861\\
\hline
\end{tabular}
\end{center}
\caption{The thresholds for 5 persons' images}
\label{THRES}
\end{table}

The dissimilarities between 3 other persons' images and 5 'known' persons' images are given in Table \ref{DISIMI}.

\begin{table}[h]
\centering
\begin{tabular}{|c|c|c|c|c|c|}
\hline
    &\multicolumn{3}{c|}{Mouth} &\multicolumn{2}{c|}{Eyes}\\
\hline
    &1  &2  &4  &3  &5\\
    \hline
\rowcolor{gray!25}
\bf Patch\_thres &\bf 0.5406 &\bf 0.5203 &\bf 1.5492 &\bf 0.4174 &\bf 0.4861\\
6\_1 &0.596875   &0.923608  &1.771417   &0.869486   &0.75737\\
\hline
6\_2 &0.709693   &0.870208  &1.681255   &1.155204   &1.031485\\
\hline
6\_3 &1.101026   &1.380079  &2.12309    &0.654594   &0.655111\\
\hline
7\_1 &0.821075   &0.94641   &1.785884   &0.922266   &0.917094\\
\hline
7\_2 &0.760753   &0.780838  &1.828294   &0.701748   &0.822338\\
\hline
7\_3 &0.589346   &0.888509  &2.171345   &0.44176    &0.785486\\
\hline
\end{tabular}
\caption{The dissimilarities between 3 other persons' images and 5 'known' persons' images}
\label{DISIMI}
\end{table}

It can be seen that the dissimilarities between three other persons' images and five 'known' persons' images are all larger than the threshold. Therefore the three other persons' can be recognized as 'unknown' rapidly.

However, instead of presenting the result of the familiarity process by 'yes' (the dissimilarity value is smaller than $thres$ ) and 'no' (the dissimilarity value is larger than $thres$), in this paper, the process result is presented by the probability of one candidate being a 'known' subject that computed by \eqref{Baye}

\begin{equation}
p(c_i|x)=\frac{p(x|c_i)p(c_i)}{\sum^n_{i=1}p(x|c_i)p(c_i)}\label{Baye}
\end{equation}

As we assume all $p(c_i)$ are the same, the identities are actually sorted by $p(x|c_i)=\exp(-d)$, where $d$ is the Euclidean distance between features of the learned face and the incoming face.

Based on \eqref{Baye}, the first five persons' images used for recognition are compared with their images used for memory.  The probability and the corresponding ranks are given in Table \ref{Prob}(only the results of 4 persons' 3 images are listed here).

\begin{table}[h]
\centering
\begin{tabular}{|c|c|>{\columncolor{gray!25}}c|>{\columncolor{gray!25}}c|>{\columncolor{gray!25}}c|c|c|}
\hline
1\_1    &Probability    &0.2836    &0.2534    &0.2207    &0.1826    &0.0597\\
\cline{2-7}
    &Ranking    &1    &5    &2    &3    &4\\
    \hline
1\_2    &Probability    &0.2975    &0.2455    &0.2207    &0.1783    &0.0579\\
\cline{2-7}
    &Ranking    &1    &5    &2    &3    &4\\
    \hline
1\_3    &Probability    &0.3413    &0.2559    &0.193    &0.1508    &0.059\\
\cline{2-7}
    &Ranking    &1    &5    &3    &2    &4\\
    \hline
2\_1    &Probability    &0.316    &0.2547    &0.2202    &0.164    &0.045\\
\cline{2-7}
    &Ranking    &2    &1    &3    &5    &4\\
    \hline
2\_2    &Probability    &0.3409    &0.2527    &0.2115    &0.1409    &0.054\\
\cline{2-7}
    &Ranking    &2    &1    &3    &5    &4\\
    \hline
2\_3    &Probability    &0.3026    &0.2474    &0.2144    &0.1819    &0.0537\\
\cline{2-7}
    &Ranking    &2    &3    &1    &5    &4\\
    \hline
3\_1    &Probability    &0.3608    &0.2202    &0.2038    &0.1838    &0.0314\\
\cline{2-7}
    &Ranking    &3    &2    &5    &1    &4\\
    \hline
3\_2    &Probability    &0.3407    &0.2198    &0.2084    &0.2023    &0.0288\\
\cline{2-7}
    &Ranking    &3    &2    &1    &5    &4\\
    \hline
3\_3    &Probability    &0.3163    &0.2341    &0.2107    &0.2099    &0.029\\
\cline{2-7}
    &Ranking    &3    &2    &1    &5    &4\\
    \hline
4\_1    &Probability    &0.4885    &0.1476    &0.1444    &0.1238    &0.0958\\
\cline{2-7}
    &Ranking    &4    &3    &2    &1    &5\\
    \hline
4\_2    &Probability    &0.4832    &0.1511    &0.1468    &0.1248    &0.0941\\
\cline{2-7}
    &Ranking    &4    &2    &3    &1    &5\\
    \hline
4\_3    &Probability    &0.4698    &0.1595    &0.1507    &0.1316    &0.0884\\
\cline{2-7}
    &Ranking    &4    &1    &3    &2    &5\\
    \hline

\end{tabular}
\caption{The probability and possibility rank of first 4 people' other images compared with their 'known' images through special patches}
\label{Prob}
\end{table}
It can be seen that each person's other images are close to themselves through familiarity matching. Three 'known' objects among top three matching probability are reserved for further recognition.

\subsection{Recognition by recollection matching}
In this process, the similarity in semantic meaning is compared.

\begin{figure}[h]
\centering
\subfloat[\label{ASMa}]{\includegraphics[width=0.3\columnwidth]{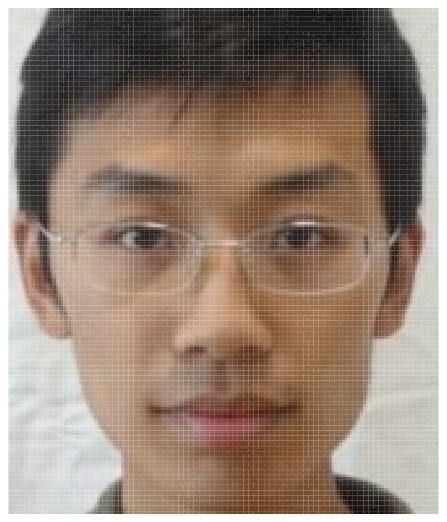}}
\subfloat[\label{ASMb}]{\includegraphics[width=0.3\columnwidth]{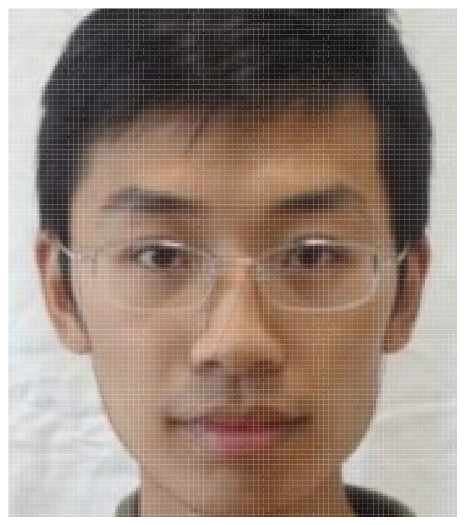}}
\subfloat[\label{ASMc}]{\includegraphics[width=0.3\columnwidth]{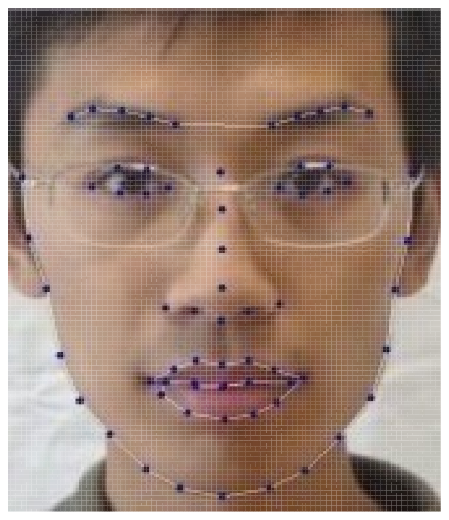}}

\caption{ (a) input image (b) compute controlled points with ASM (c) affine transformation with manual modifying.}
\label{ASM}
\end{figure}

In this paper, the controlled points of faces are obtained by the Active Shape Model (ASM) or manually.  One sample (68 control points can be obtained from 1632*1632 image) is given in Fig. \ref{ASM}.

The main parts of a face include eyes, nose and mouths. The semantic presentation has 15 attributes as shown in Section \ref{SectionSemantic}. Therefore, one 'known' face is saved by 15 semantic attributes, 1 prominent attribute which determines the part of special patch, 1 special patch and thresholds (See Table \ref{ATTR}).

\begin{table}
\centering
\begin{tabular}{|c|c|}
\hline
Features                &Dimensions\\
    \hline
Geometric attributes    &15\\
    \hline
Prominent attributes    &1\\
    \hline
Special patch           &scale $[4, 6, 8, 10]\times8$\\
    \hline
Attributes similarity threshold &1\\
\hline
C2 similarity threshold         &1\\
\hline

\end{tabular}
\caption{Attributes for one 'known' face.}
\label{ATTR}
\end{table}

The attribute thresholds computed by function 4-1 for 5 persons are shown in following Table \ref{Thresholds}:

\begin{table}
\centering
\begin{tabular}{|c|c|c|c|c|c|}
\hline
identity    &1  &2  &4  &3  &5\\
\hline
attri\_thres    &4.5946 &4.1696 &4.0039 &4.4526 &3.5100\\
\hline
\end{tabular}
\caption{The thresholds for attributes}
\label{Thresholds}
\end{table}

By computing the difference between the test image and the three candidate persons' attributes in turn, and compared with the attributes thresholds. We could give the final label for each test image.

The recognition results are listed in Table \ref{Result}.
\begin{table*}
\begin{tabular}{|c|c|c|c|c|c|c|c|c|c|c|}
\hline
identity    &\multicolumn{2}{c|}{1}    &\multicolumn{2}{c|}{2}    &\multicolumn{2}{c|}{3}    &\multicolumn{2}{c|}{4}    &\multicolumn{2}{c|}{5}\\
\hline
attri\_thres &\multicolumn{2}{c|}{4.5946}    &\multicolumn{2}{c|}{4.1696}    &\multicolumn{2}{c|}{4.0039}    &\multicolumn{2}{c|}{4.4526}    &\multicolumn{2}{c|}{3.5100}\\
\hline

Test	&label	&Sim	&label	&Sim	&label	&Sim	&label	&Sim	&label	&Sim\\
\hline
X\_1	&1	&2.1773	&2	&2.0454	&3	&3.2711	&4	&1.9933	&5	&2.1834\\
\hline
X\_2	&1	&2.6458	&2	&1.7975	&3	&2.9305	&4	&1.8047	&5	&2.2325\\
\hline
X\_3	&1	&2.3152	&2	&2.507	&3	&2.1742	&4	&3.8357	&5	&3.3121\\
\hline
X\_4	&1	&2.332	&2	&2.0656	&3	&3.7265	&4	&3.4721	&*	&$\backslash$\\
\hline
X\_5	&1	&3.2519	&2	&3.9128	&3	&2.6567	&4	&1.6642	&3	&3.5643\\
\hline
X\_6	&0	&$\backslash$	&2	&1.5454	&3	&3.1565	&4	&2.3439	&5	&2.5769\\
\hline
X\_7	&1	&3.5434	&2	&1.8358	&3	&2.219	&4	&2.9189	&*	&$\backslash$\\
\hline
X\_8	&1	&3.1574	&2	&2.317	&0	&$\backslash$	&4	&3.1242	&*	&$\backslash$\\
\hline
\end{tabular}
\caption{Recognition results. '*' indicates that the person is wrongly rejected during familiarity discrimination. '0' indicates that the person is wrongly rejected during recollection matching, and '$\backslash$' means that the similarity is missing since the person is rejected in previous process.}
\label{Result}
\end{table*}


\section{Conclusion}
In this paper, the biologically inspired memory and association model has been proposed.  The typical features of the model based on biological work include:

\begin{enumerate}[(a)]
\item
in both the memory and recognition processes, the objects have semantic and episodic features and episodic patch is extracted according to semantic feature.
\item
The sensitivities to the semantic and episodic features are learnt during the process.
\item
In association process, the semantic and episodic features are separately compared and the results are integrated, which correspond to familiarity and recognitive matching.
\end{enumerate}

Through six blocks, the above model is introduced to the HMAX and the corresponding algorithms are given.  The memory and association mechanism provides 'top-down' control function in the process, which can active select the important features in memory and association and influence the recognition process.

The experimental results show that with the new method, the required memory is very small and recognition performance is very good.

In the future work, the combined model would be further improved with more learning ability.

%
%
%
\section*{Acknowledgment}

The authors would gratefully thank Dr. Wei Wu and Dr. Tianshi Chen for their kindly help to improve this paper.

\ifCLASSOPTIONcaptionsoff
  \newpage
\fi




\bibliographystyle{IEEETran}
\bibliography{SMCB_paper}
%
%
%
%
%

\end{document}